# Precision Sugarcane Monitoring Using SVM Classifier

Sachin Kumar[a], Sumita Mishra[a], Pooja Khanna[b], Pragya[c]
*a (Department of Electronics and communication Engineering, Amity University, Lucknow skumar3@lko.amity.edu, smishra@lko.amity.edu)*
*b (Department of Computer Science Engineering, Amity University, Lucknow Campus pkhanna@lko.amity.edu)*
*c (Department of Chemistry, MVPG College, Lucknow, dr.pragya2011@gmai.com)*

**Abstract**

**India is agriculture based economy and sugarcane is one of the major crops produced in northern India. Productivity of sugarcane decreases due to inappropriate soil conditions and infections caused by various types of diseases ; timely and accurate disease diagnosis, plays an important role towards optimizing crop yield. This paper presents a system model for monitoring of sugarcane crop, the proposed model continuously monitor parameters (temperature, humidity and moisture) responsible for healthy growth of the crop in addition KNN clustering along with SVM classifier is utilized for infection identification if any through images obtained at regular intervals. The data has been transmitted wirelessly from the site to the control unit. Model achieves an accuracy of 96% on a sample of 200 images; the model was tested at Lolai, near Malhaur, Gomti Nagar Extension.**

## 1. Introduction

Food production in India is at present heavily reliant on Agriculture. [1].. Sugercane is one of the major crops in India as Sugarcane shares 7% of the total agriculture production in India [3]. Crop diseases are a potential threat to adequate crop yield and it has disastrous consequences for farmers ; especially those with small farm area, whose livelihoods depend on healthy crops[2].

Precision agriculture is one of emerging field in engineering that utilizes information technology image processing techniques to optimize the agricultural cultivation process. significant features of PA are cost reduction towards identification of crop diseases ;subsequently application of these techniques leads to fast and accurate diagnosis of crop disease..It facilitates the timely provision of the valuable information for the prevention and treatment of diseases resulting in improved crop yield precision agriculture aims to accurately analyze parameters such as the temperature, humidity, soil condition in order to extract timely information to maximize crop yield.

we have designed a system for sugercane agriculture management, which based on various sensor inputs combined with crop images to identify cultivation environment and infections in the crop to suggest varoious measures that may be useful towards optimizing the crop yield. The system utilizes sensor nodes that collect and transmit data about the quality of the water supply, the soil, and temperature in an agricultural field. While such sensor-based systems have been investigated earlier, one of the unique features in our system is the combination of these sensors systems with an image processing model to realize a comprehensive system for management of sugarcane cultivation.

The paper is organized as follows: in part II we describe the System model used. Part III presents the implementation of the proposed model, Part IV presents result obtained, Finally, conclusion is presented.

## 2. Proposed System Model

System architecture proposed is shown in figure 1, proposed model is broadly divided in to Control Unit, Data Acquisition (Images and sensor data) and Processing Unit.

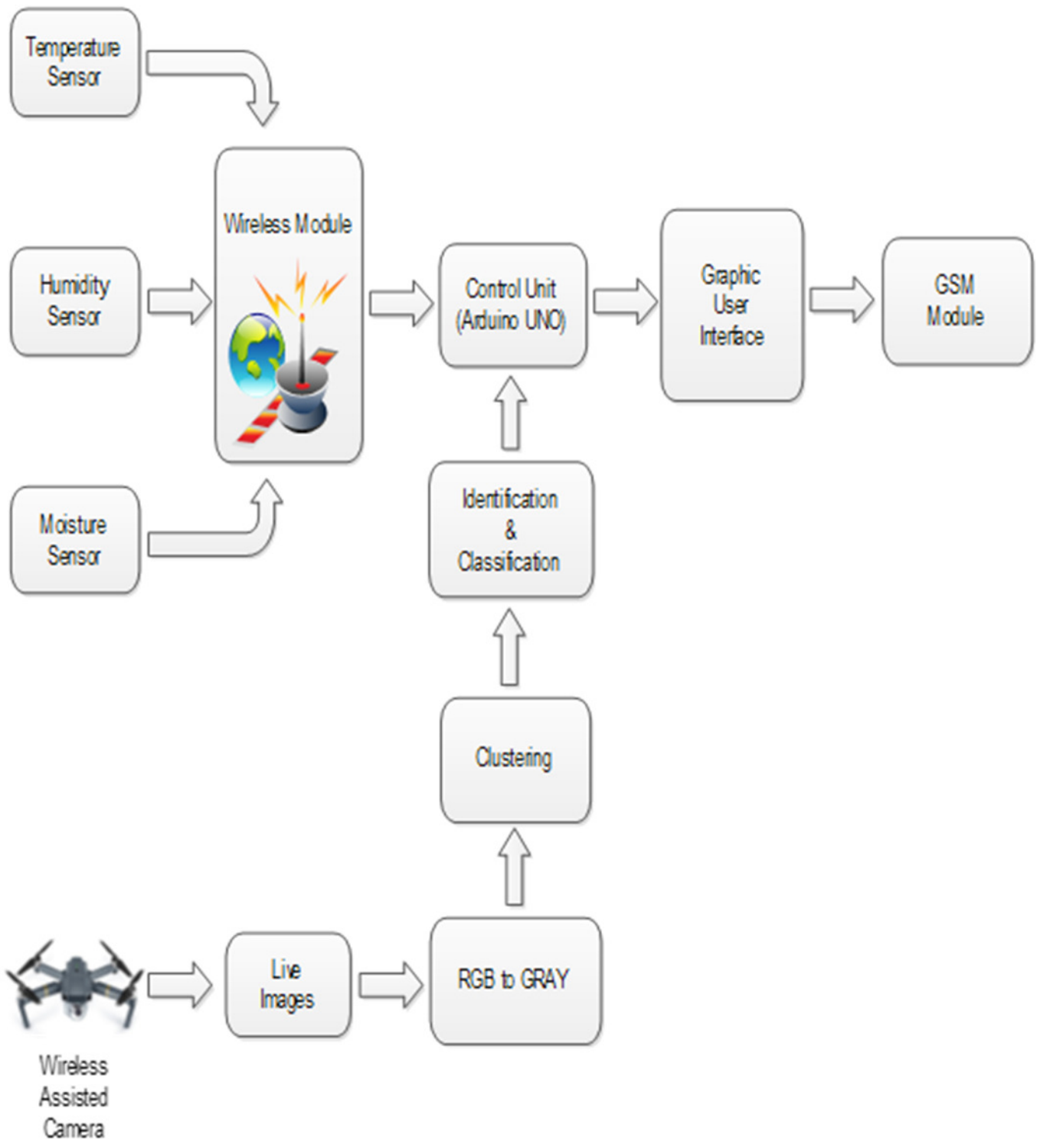

Figure 1. Proposed System Model

Sensors (temperature, moisture and humidity) have been utilized to monitor and maintain favorable conditions for crop growth and Classification Algorithms in sequence with Images Processing algorithm has been used to detect and identify infections and diseases if any in crop. The data so gathered is being fed to Control Unit where in, any deviation from fruitful growth is monitored and subsequently necessary steps to improve is being suggested to the farmer.

## 3. Implementation

The architecture consists of data acquisition and its analysis as its primary components, further deductions from analysis decides the suggestive course of action to be adopted. Acquisition involves data from sensors and Images from camera. Sensors used are temperature, Humidity and Moisture

**Temperature Sensors:** Favorable temperature conditions are one of the key factors for healthy growth of the crop. Ideal temperature for germination is 32° to 38°C. Growth rate reduces below 25°, becomes most favorable at about 30°-34°, growth rate again slows down beyond 35° and almost becomes stagnant beyond 38°.Above 38° photosynthesis slows down and respiration increases. Temperature in between 12° to 14° are desirable for ripening, thus temperature plays pivotal role in the fruitful growth of the plant, therefore monitoring of temperature via a suitable sensor is extremely important.

**Humidity Sensors:** High content of humidity (80 -85%) provides ideal condition for sugarcane growth, especially in early part of growth period, therefore an optimum value of 50-65% in tandem with limited water supply is perfect for ripening. Soil moisture sensors (SMS) provide one of the best and simple solutions to this requirement, thus help in on farm water management decisions.

**Moisture Sensor:** Moisture sensors typically estimate water content present in soil by volume. Evaluation of amount of moisture in soil is extremely important for farmers, as the irrigation systems needs to be regularized. This helps in increasing productivity and quality of crop, further water loss also reduces.

Measurement of Temperature and Humidity for the sugarcane crop, DHT11 has been used, depicted in figure 2. DHT sensors have two different functional blocks a thermistor for temperature measurement and a capacitive humidity sensor. Sensor comprises of an embedded analog to digital converter that outputs temperature and humidity in digital format. Sensor refreshes its data after every 2 seconds, so readings can only be up to 2 seconds old.

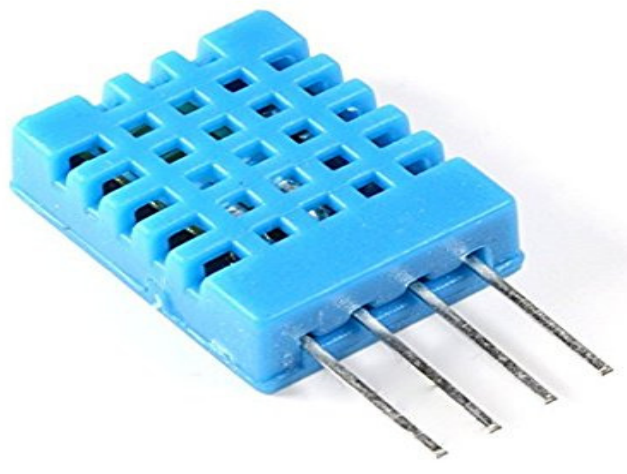

Figure 2: DHT11 Sensor

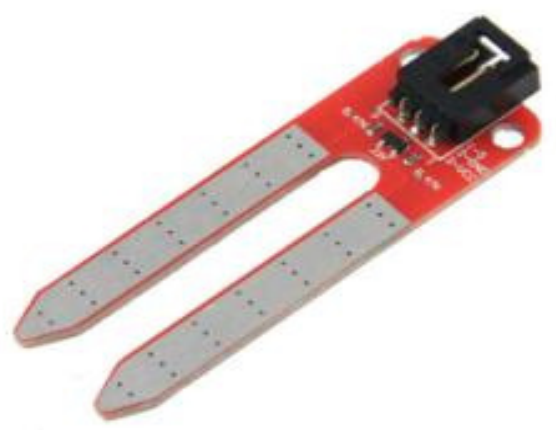

Figure 3: Moisture Sensor Probe

Moisture Sensor uses the principle of difference in resistance offered between dry and moist soil in between the probes on conduction of current to evaluate the moisture level of the soil, Moisture sensor utilized for the purpose is shown in figure 3. [4-6]

Images at regular intervals during the crop growth, from camera permanently planted in farm have been utilized for monitoring of infections. There are various leaf diseases in sugarcane. Some Visible leaf diseases which can be detected by the image processing techniques are:

- **Disease : Leaf scald:**

The infection is latent and therefore it can remain unobserved and unseen for quite some time, by the time infection becomes visible, the crop might get seriously infected.
Primary symptoms include development of white lines like structures with yellow outline following veins of the leaf that causes death of the infected tissue; these are known as “pencil lines”. The infection causes leaf to lose their original color to become pale green in color as they fail to generate chloroplast. Figure 4 depicts Leaf scald disease.

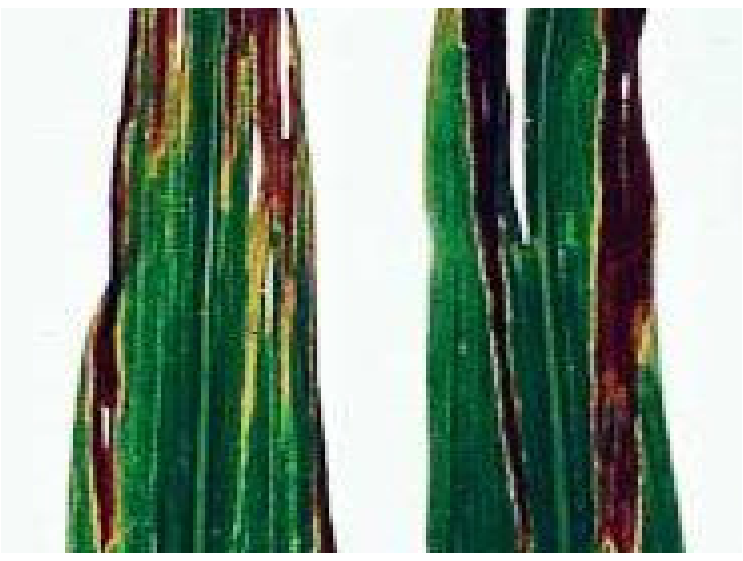

Figure 4: Leaf Scald Disease Survival and s

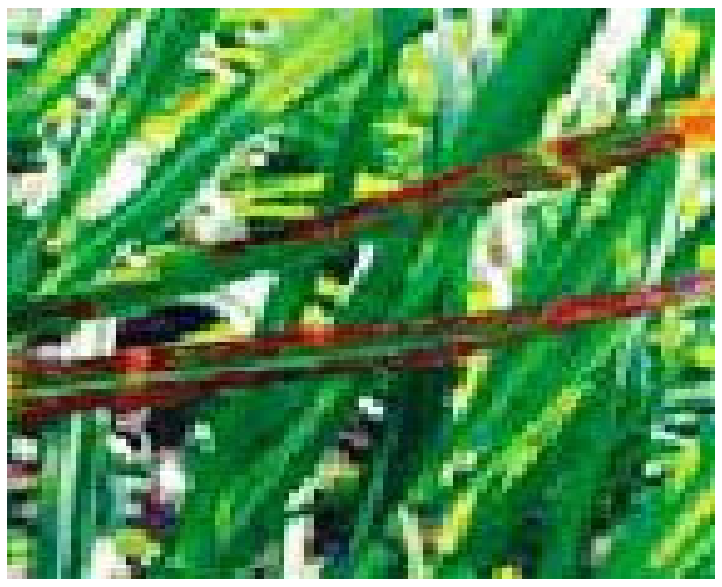

Figure 4: Red Striped Disease Survival and spread

Disease usually spread through Pathogens that survive in cane stubble, grass and on agricultural implements. Severity of the disease increases with periods of stress, like waterlogging, low temperature and drought.

- **Disease : Red striped**

Disease is characterized by the presence of stripes red in color , 0.5 – 1.00 millimeter wide and several millimeter in length, either concentrated in center or spread all over the blade. Red stripes are chlorotic lesions, usually several coalesce together covering entire surface of the leaf blade causing wilting and drying of the leaves. Flakes white in color occur on lower portion of the leaf, these are dry bacterial ooze. Young shoot yellow in color, when affected have reddish brown stripes appear on shoots. On detection plants are cut by dissecting the shoot downward, where red colored discoloration of tissues may be observed

In the affected crop vascular bundles are distinguished due to the dark red discoloration, as depicted in figure 5.

Disease can spread through wind and rain, Humid conditions are ideal for such bacterial infection.

- **Disease: Mosaic**

Disease is detected through appearance of Mosaic pattern, depicted in figure 6 , usually detected when leaves of the crown are held against the bright light source. Area affected displays reddening or necrosis. Sometime leaf sheath may also display similar patterns.

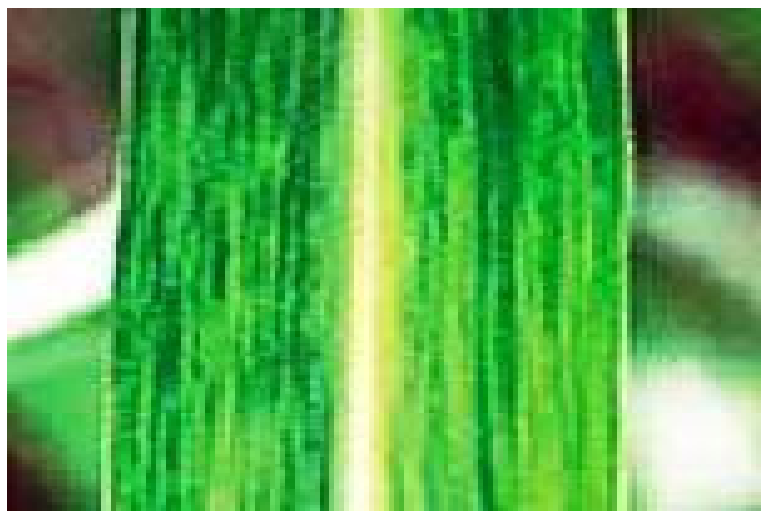

Figure 6: Mosaic Disease

## Image Acquisition

Infected sugercane leaves with typical symptoms of Leaf scald disease, Red striped disease and mosaic disease used in this study were sampled from the sugarcane plantation, Lolai, near Malhaur, Gomti Nagar Extension, Lucknow.

For infection identification, image processing algorithms have been used, for ease of processing images are operated in Gray scale, segregation of infected area was achieved by K-mean clustering further segmentation was obtained by morphological operations. The area segmented was then subjected to classification using SVM classifier for infection identification based on specific features related to various infections. [7,8]
Image classification problem is addressed in two steps first clustering based unsupervised algorithm, K-mean is used to perform soft classification which retains more information and identifies region of interest. Features used include color and shape infection. Final classification problem is restricted as the two-class problem with the classes being healthy and infected. Segregation of infected region and elimination of isolated pixels is achieved via morphological operations , opening given by equation (ii) and depicted in figure 3(c)

$$A \circ B = (A \oplus B) \oplus B \text{.........ii}$$

Opening performs dilation followed erosion function on the identified infected region A, B denotes structuring element utilized for performing erosion and dilation operation. Opening operation segregates infected area as outer line of the infection. Depending upon the quality of area segmented, few clinical processes like skeletization are required to further outline infected region to a finer level. The noise is further eliminated be the help of median filter.

$$Perimeter = \sum A_{EDGES} B_{EDGES}\, C_{EDGES}\, D_{EDGES} \text{ ...iii}$$

Where $A_{EDGES}$, $B_{EDGES}$, $C_{EDGES}$ and $D_{EDGES}$ are top, left, bottom and Right corners of the edges of the perimeter being evaluated. Clustered data after refinement is further classified using SVM classifier. A set of training images ($x_1$, $y_1$), ($x_2$, $y_2$), ..., (xn, yn); where xi is input with d attributes

$$x_i \mathcal{E}\, R^d$$

$$y_i \mathcal{E}\{-1, +1\};\ \text{i} = 1{,}2$$

SVM creates the optimal separating hyper Plane separating two classes of data by using a kernel function (K). hyper plane was based on features obtained from clustering (Color and shape), this hyperplane is represented as

$$\text{w} \bullet \text{x} + \text{b} = 0,$$

*where*: *w* represents the normal to the hyperplane and *b* represents the perpendicular distance from hyperplane to origin
All images for which feature vector lies on one side of the hyper plane, belong to class +1, labeled as infected

$$\text{w} \cdot \text{x i} + \text{b} \geq +1 \text{ for yi} = +1$$

and the images for which feature vector satisfies eqn 2 belong to class -1, labeled as benign.

$$w \cdot x_i + b \leq +1 \text{ for } y_i = -1$$

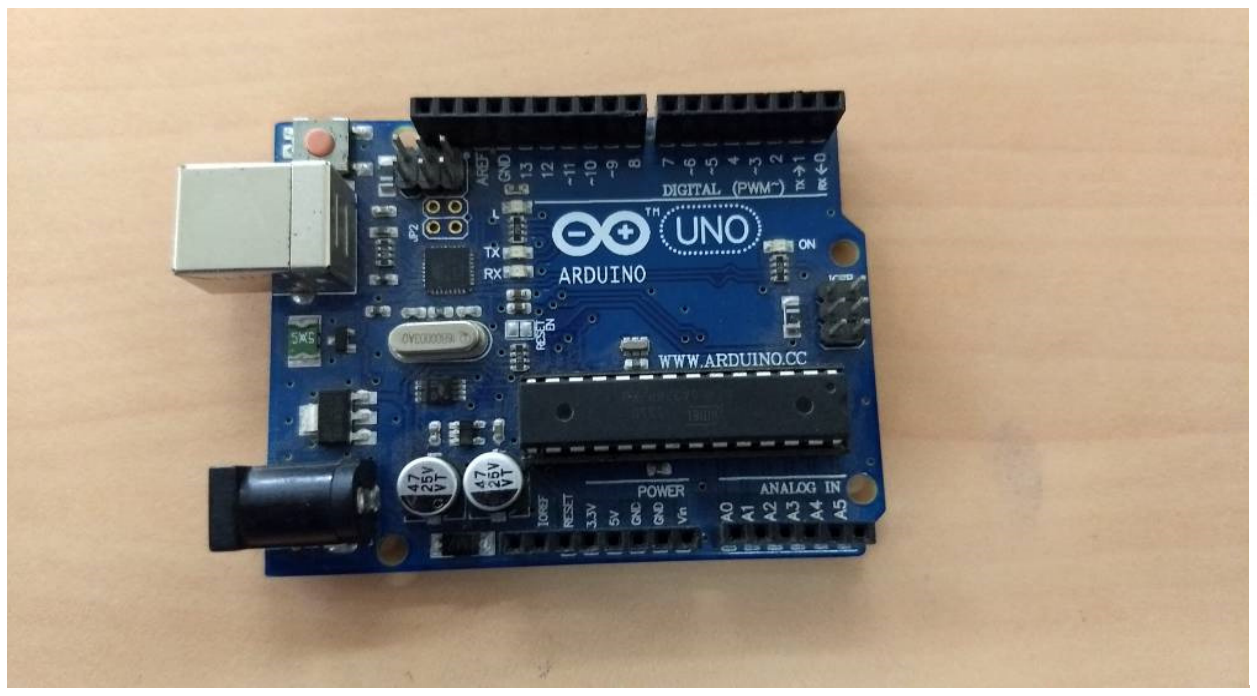

Figure 6: Arduino UNO

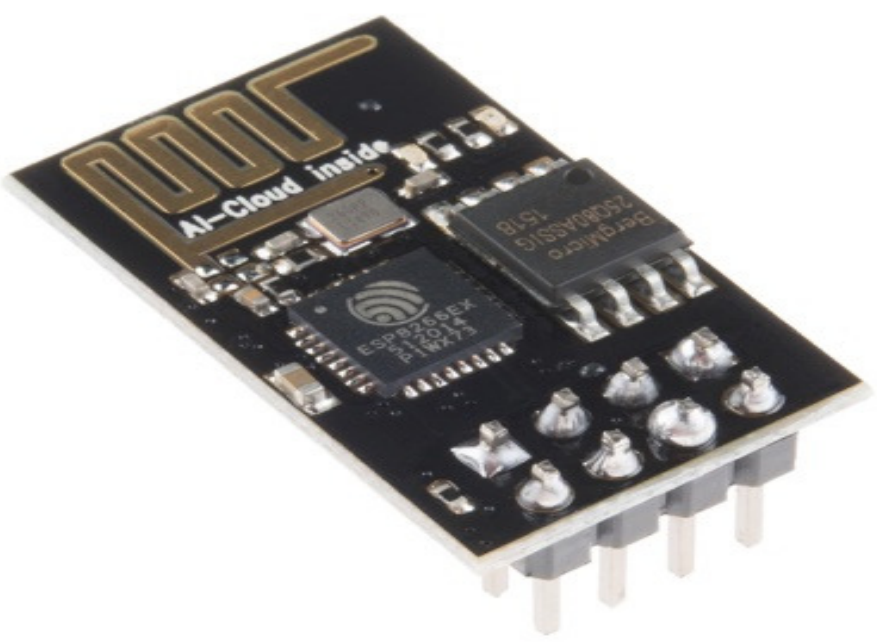

Control Unit is in the form of Arduino UNO, Arduino UNO comprises of microcontroller ATmega328P, as depicted in figure 7 embedded in the board. Arduino has a Software platform which is open source, which simplifies the task of building and executing the code. Software is compatible with Linux, Mac and Windows Software platform is developed on JAVA environment. Board is programmed to make deductions on the basis of inputs received from sensors and classifier algorithm via wireless module (ESP8266 Wi-Fi Module). ESP8266 is a self-contained IC embedded with TCP/IP protocol, for convenient access to Wi-Fi network. ESP8266 is depicted in figure 7, is capable of hosting an application and offloading all Wi-Fi functions from a remote application microcontroller. [9-13]

Figure 7: Wi-Fi Module (esp8266)

GUI for the model has been designed on Matlab platform depicting sensor and classifier output. Proposed system has implemented the alert system to the farmer using GSM module **(SIM800).** SIM800 supports Quad-band 850/900/1800/1900 Mega Hertz, it can wirelessly transmit data, Voice and SMS using extremely low power, shown in figure 8. With tiny size of 24*24*3mm, it can almost fit into any space and thus can meet any demands of customer design. Loaded with Bluetooth and Embedded AT, it helps in cost cutting and lesser delays in time-to-market for customer applications. Figure 9 depicts the complete circuit design for implementation using EAGLE 7.1.O

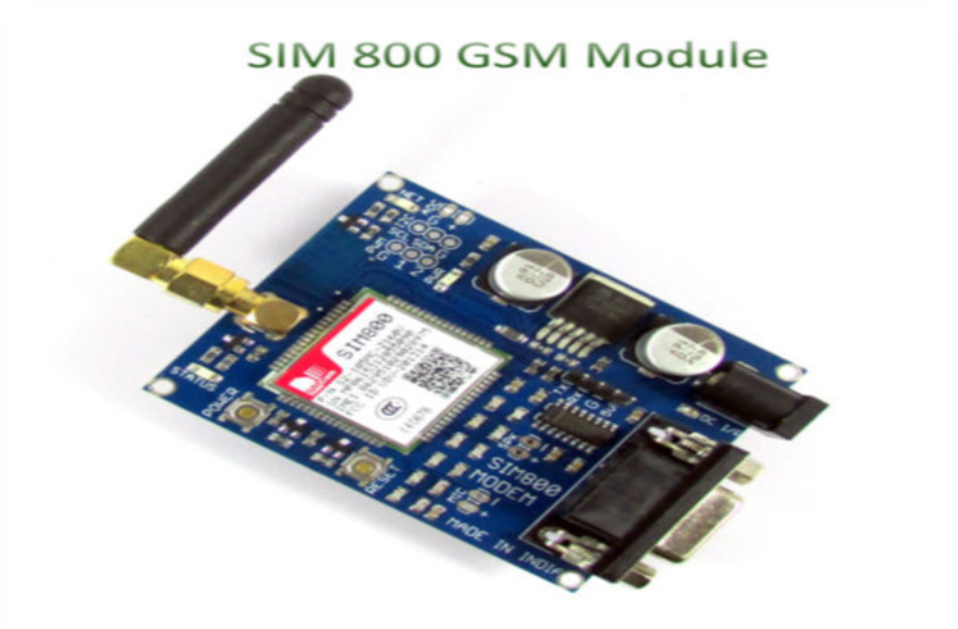

Figure 8: GSM Module

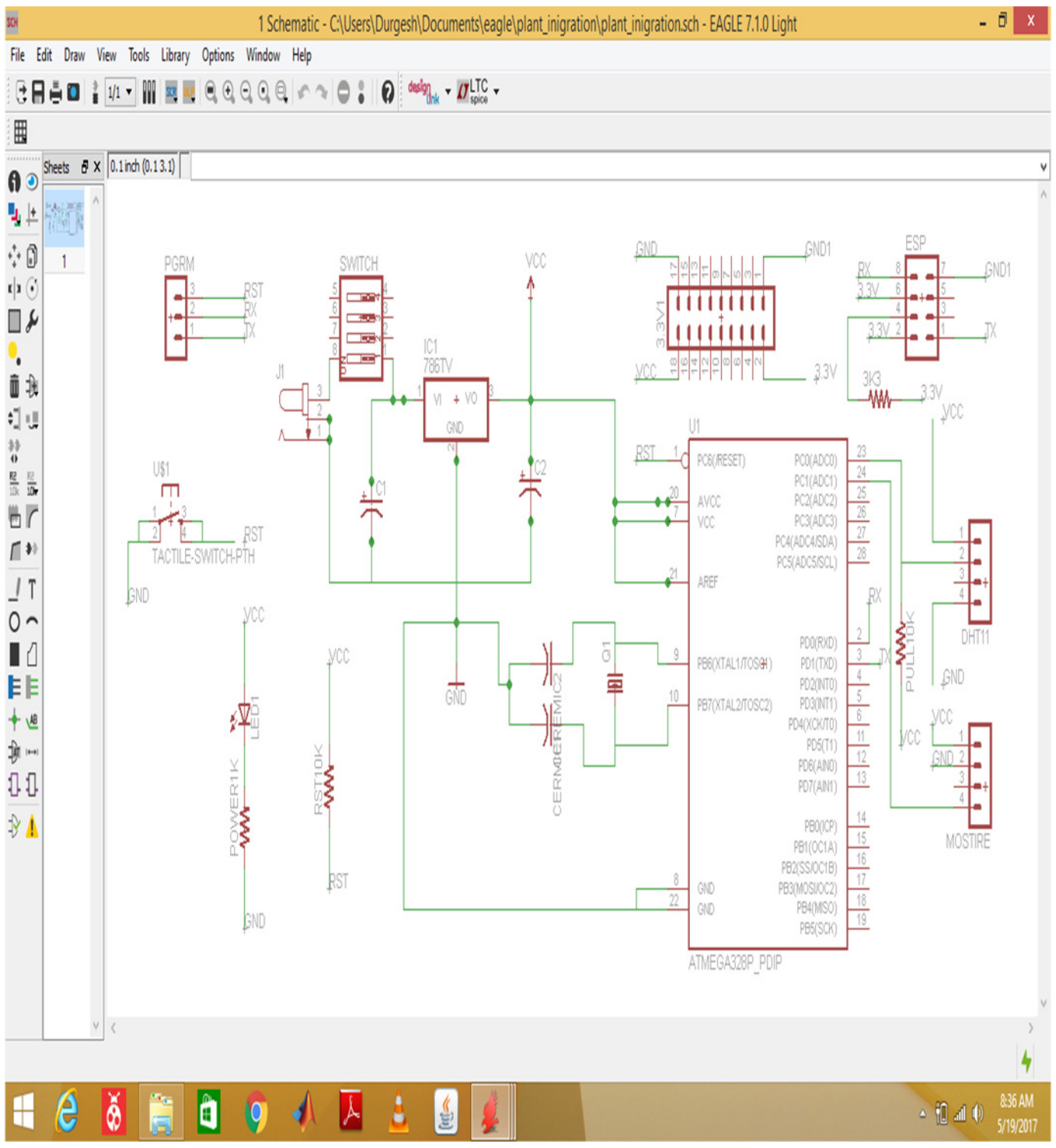

Figure 9 depicts the complete circuit design for implementation using EAGLE 7.1.O

**Result**

System model was tested at Lolai, near Malhaur, Gomti Nagar Extension, Lucknow. Sensors and Camera were deployed at the site where sugarcane was sown. Images and data was transmitted wirelessly to the Control Unit, where in further analysis was done. Control Unit on the basis of analysis provides suggestive measures to the farmers. The data generated is shown on GUI , where ideal values of various sensors are pre-defined. The measured values are compared with the ideal values. If there are changes in the measured values then there is an Automatic message facility in the GUI. With the help of GSM module an automatic message is send to the registered number. Graphic User Interface display graphs for temperature, Humidity and Moisture values, and further it also displays leaf section to monitor infections with the help of image processing and classification algorithms. System was tested for 200 sample images and accuracy of 96% was achieved

Figure 10a depicts designed GUI with Mosaic disease identification, Figure 10b depicts designed GUI with Leaf Scald disease identification, Figure 10c depicts designed GUI with Eye Spot disease identification and Figure 10d depicts designed GUI with healthy leaf.

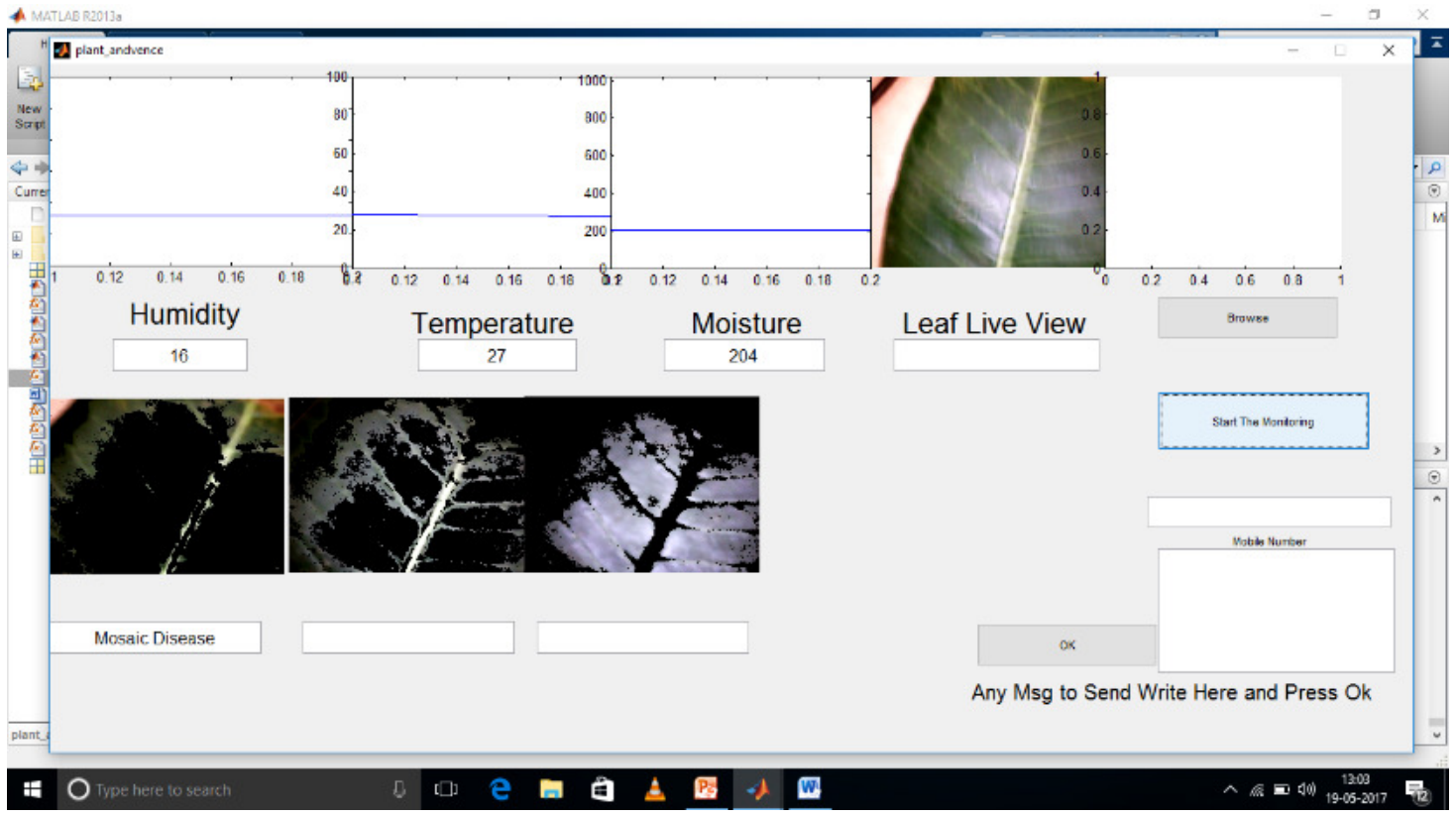

Figure 10a GUI with Mosaic disease identification

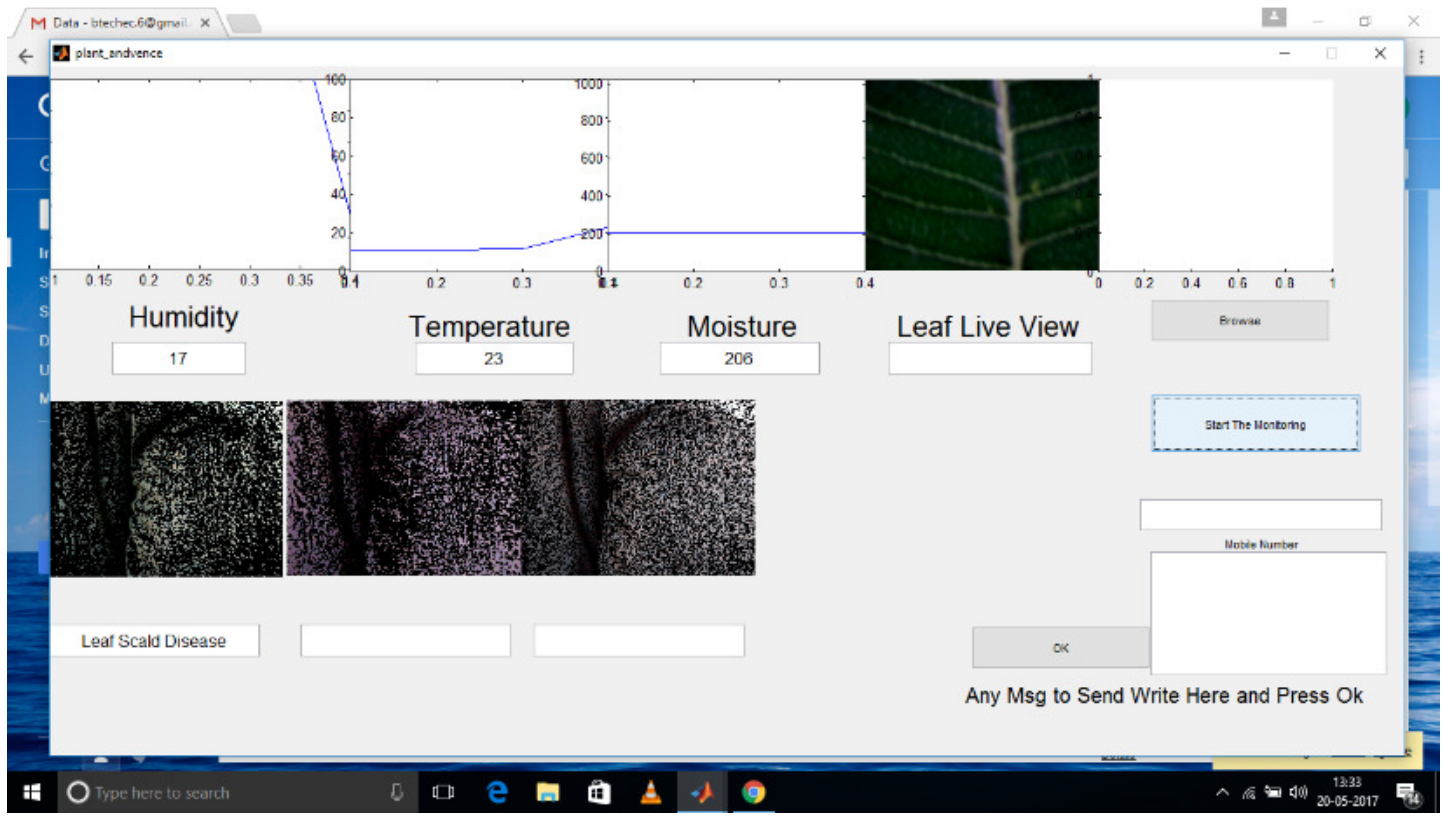

Figure 10b GUI with Leaf Scald disease identification

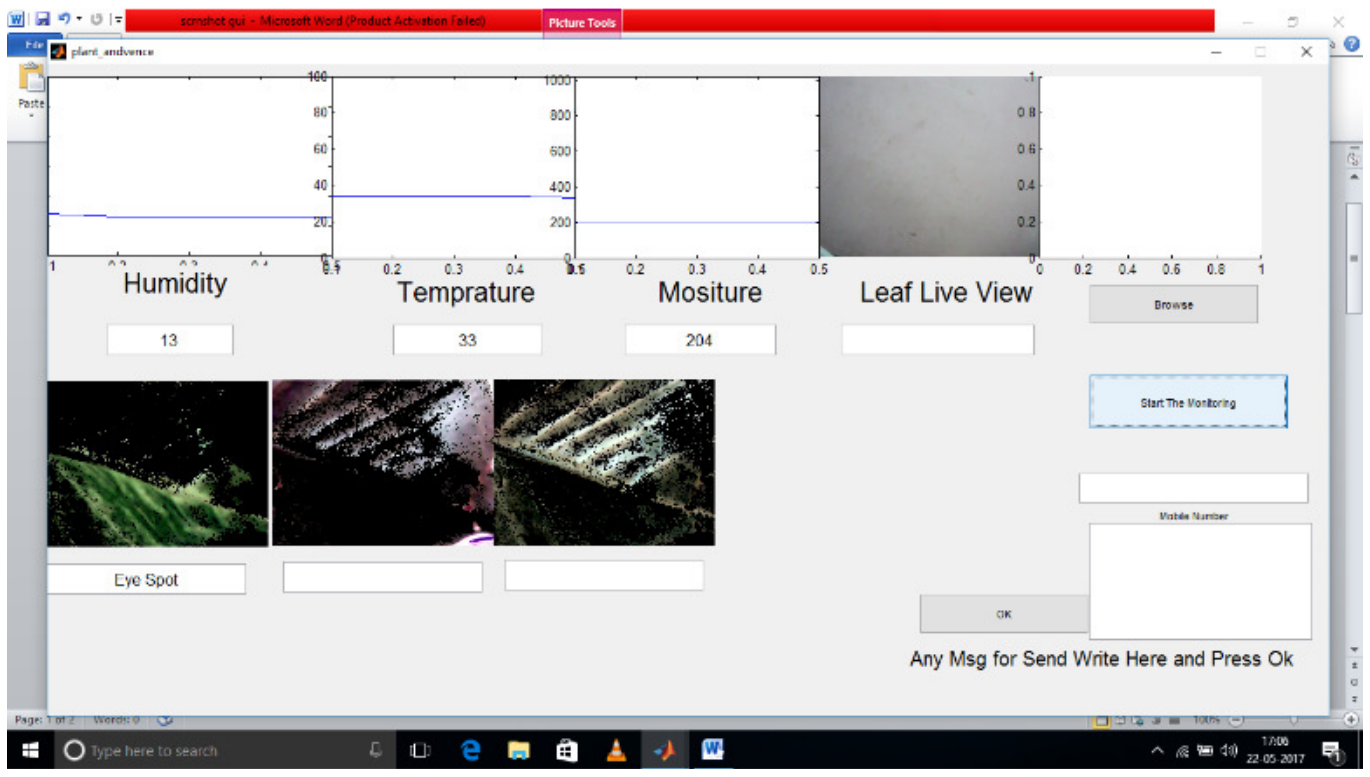

Figure 10c GUI with Eye Spot  disease identification

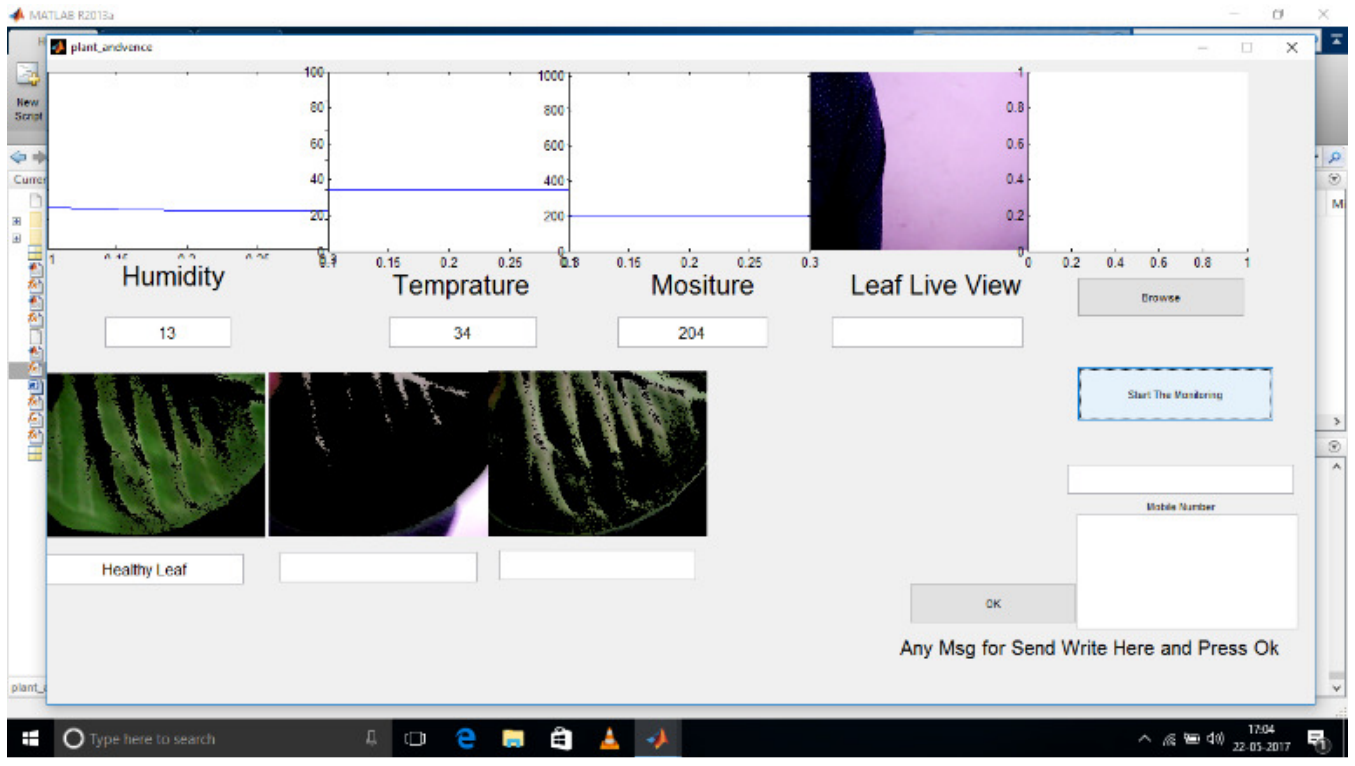

Figure 12d GUI with healthy leaf

## Conclusion

Agriculture forms the backbone of most of the developing countries; Precision agriculture methods with smart algorithms are the need of hour to meet current and future requirements of crops.This Paper presents a model for smart agriculture technology to limit resources and to increase productivity of sugarcane in safe and secure environment. Proposed model continuously monitor parameters (temperature, humidity and moisture) responsible for healthy growth of the crop in addition KNN clustering along with SVM classifier is utilized for infection identification if any through images obtained at regular intervals , accuracy achieved is 96% on sample of 200 images, the model was tested at Lolai, near Malhaur, Gomti Nagar Extension. System model presented integrates ideal parameter monitoring through wireless sensor network along with image processing and classification algorithms for early and automated detection of diseases that might infect the crop.